\documentclass[a4,a4wide]{article}
\usepackage{aaai}
\usepackage{times}
\usepackage{helvet}
\usepackage{courier}
\usepackage{url}
\usepackage{tikz}
\usepackage{float}
\usepackage{stmaryrd}
\usepackage{amssymb}
\usepackage{amsmath}
\usepackage{amsthm}
\usepackage{graphicx}
\usepackage{float}
\usepackage{todonotes}
\usepackage{nameref}
\usepackage{paralist}
\usepackage{multirow}
\usepackage{soul}
\usepackage{subcaption}
\usepackage{thmtools}

\title{Extension--based Semantics of Abstract Dialectical Frameworks}

\author{Sylwia Polberg \\ Vienna University of Technology \\Institute of Information Systems\\ Favoritenstra\ss{}e 9-11, 1040 Vienna, Austria
\thanks{The author is funded by the Vienna PhD School of Informatics. This
research is a part of the project I1102 supported by the Austrian Science Fund FWF.}}

\newtheorem{theorem}{Theorem}[section]
\newtheorem{lemma}[theorem]{Lemma}

\newtheorem{definition}[theorem]{Definition}
\newtheorem{proposition}[theorem]{Proposition}

\declaretheorem[style=plain, sibling=theorem]{example}
\renewcommand\thmcontinues[1]{Continued}

\theoremstyle{remark}

   {\begin{trivlist}\item[]\textit{Proof Idea.}}%
   {\end{trivlist}}

   {\begin{trivlist}\item[]\textit{Proof Sketch.}}%
   {\end{trivlist}}

\newcommand{\tvt}{\mathbf{t}}
\newcommand{\tvf}{\mathbf{f}}
\newcommand{\tvu}{\mathbf{u}}
\begin{document}
\nocopyright
\maketitle

\begin{abstract}
One of the most prominent tools for abstract argumentation is the Dung's framework, AF for short.
It is accompanied by a variety of semantics including grounded, complete, preferred and stable. Although powerful, AFs have their shortcomings, which led to
development of numerous enrichments. Among the most general ones are the abstract dialectical frameworks, also known as the ADFs.
They make use of the so--called acceptance conditions to represent arbitrary relations. This level of abstraction brings not only new
challenges, but also requires addressing existing problems in the field. One of the most controversial issues, recognized not
only in argumentation, concerns the support cycles. In this paper we introduce a new method to ensure acyclicity of the chosen arguments and present a family
of extension--based semantics built on it. We also continue our research on the semantics that permit cycles and fill in the gaps from the previous works.
Moreover, we provide ADF versions of the properties known from the Dung setting.
Finally, we also introduce a classification of the developed sub--semantics and relate them to the existing labeling--based approaches.
\end{abstract}

\section{Introduction}

Over the last years, argumentation has become an influential subfield of artificial intelligence \cite{book:argai}.
One of its subareas is the \textit{abstract argumentation}, which became especially popular thanks to the research of
Phan Minh Dung \cite{article:dung}. Although the framework he has developed was
relatively limited, as it took into account only the conflict
relation between the arguments, it inspired a search for more general models (see \cite{general} for an overview). Among the most abstract enrichments
 are the abstract dialectical frameworks, ADFs for short \cite{inproc:adf}. They make use of the so--called
acceptance conditions to express arbitrary interactions between the arguments. However, a framework cannot be 
considered a suitable argumentation tool without properly developed semantics. 

The semantics of a framework are meant to represent what is considered rational.
Given many of the advanced semantics, such as grounded or complete,
we can observe that they return same results when faced with simple, tree--like frameworks.
The differences between them become more visible when we work with more complicated cases. 
On various occasions examples were found for which
none of the available semantics returned satisfactory answers. This gave rise to new concepts:
for example, for handling indirect attacks and defenses we have prudent and careful semantics \cite{inproc:careful,inproc:prudent}.
For the problem of even and odd attack cycles we can resort to some of the SCC--recursive semantics \cite{BaroniGG05a}, while for treatment of self attackers, sustainable 
and tolerant semantics were developed \cite{Bodanza2009}. 
Introducing a new type of relation, such as support, creates additional problems. 

The most controversial issue in the bipolar setting concerns the support cycles and is handled differently from
formalism to formalism. Among the best known structures are the Bipolar Argumentation Frameworks (BAFs for short) \cite{incoll:bipolar,article:newbaf},
Argumentation Frameworks with Necessities (AFNs) \cite{incoll:newafn} and Evidential Argumentation Systems (EASs) \cite{inproc:eas}.
While AFNs and EASs  discard support cycles, BAFs do not make such restrictions. In ADFs cycles are permitted
unless the intuition of a given semantics is clearly against it, for example in stable and grounded cases.
This variety is not an error in any of the structures; it is caused by the fact that, in 
a setting that allows more types of relations, a standard Dung semantics can be extended in several ways. 
Moreover, since one can find arguments both for and against any of the cycle treatments, lack of consensus as to what approach is the best
should not be surprising.

Many properties of the available semantics can be seen as "inside" ones, i.e. "what can I consider rational?". On the other hand,
some can be understood as on the "outside", e.g. "what can be considered a valid attacker, what should I defend from?". 
Various examples of such behavior exist even in the Dung setting. An admissible extension is conflict--free and defends against attacks
carried out by any other argument in the framework. We can then add new restrictions by saying that self--attackers are not rational. Consequently,
 we limit the set of arguments we have to protect our choice from. In a bipolar setting, we can again define admissibility in the basic manner. However,
one often demands that the extension is free from support cycles and that we only defend from acyclic arguments,
thus again trimming the set of attackers. From this perspective semantics can be seen as a two--person discussion, describing what "I can claim" and "what my
opponent can claim". This is also the point of view that we follow in this paper. Please note that this sort of dialogue perspective can already be found in argumentation
\cite{inproc:dungdialect,inproc:dialect}, although it is used in a slightly different context.

Although various extension--based semantics for ADFs have already
been proposed in the original paper \cite{inproc:adf}, many of them were defined only for a particular ADF subclass called the bipolar and
were not suitable for all types of situations. As a result, only three of them -- conflict--free, model and grounded -- remain. 
Moreover, the original formulations did not solve the problem of positive dependency cycles. Unfortunately, neither did
the more recent work into labeling--based semantics \cite{tofix:newadf}, even though they solve most of the problems of their predecessors.
The aim of this paper is to address the issue of cycles and the lack of properly developed extension--based semantics. 
We introduce a family of such semantics and specialize them to handle
the problem of support cycles, as their treatment seems to be the biggest difference among the available frameworks. 
Furthermore, a classification of our sub--semantics in the inside--outside fashion that we have described before is introduced. We also recall
our previous research on admissibility in \cite{inproc:adm} and show how it fits into the new system.
Our results also include which known properties, such as Fundamental Lemma, carry over from the Dung framework. 
Finally, we provide an analysis of similarities and differences between
the extension and labeling--based semantics in the context of produced extensions.

The paper is structured as follows. In Sections \ref{sec:dungintro} to \ref{sec:adf} we provide a background on argumentation frameworks.
Then we introduce the new extension--based semantics and analyze their behavior in Section \ref{sec:sem}. 
We close the paper with a comparison between the new concepts and the existing
labeling--based approach. 

\section{Dung's Argumentation Frameworks}
\label{sec:dungintro}

Let us recall the abstract argumentation framework by Dung \cite{article:dung} and its semantics. 
For more details we refer the reader to \cite{article:semintro}.

\begin{definition}
A \textbf{Dung's abstract argumentation framework} (AF for short) is a pair $(A, R)$, where $A$ is a set of arguments and
$R \subseteq A \times A$ represents an attack relation.
\end{definition}
\begin{definition}
Let $AF = (A, R)$ be a Dung's framework. We say that an argument $a \in A$ is \textbf{defended}\footnote{Please note defense is often
also termed acceptability, i.e. if a set defends an argument, the argument is acceptable w.r.t. this set.}
 by a set $E$ in $AF$, if for 
each $b \in A$ s.t. $(b, a)\in R$, there exists $c \in E$ s.t. $(c, b)\in R$. A set $E \subseteq A$ is:
\begin{itemize}
\item \textbf{conflict--free} in $AF$ iff for each $a, b \in E,\, (a,b) \notin R$. 
\item \textbf{admissible} iff conflict--free and defends all of its members.
\item \textbf{preferred} iff it is maximal w.r.t set inclusion admissible. 
\item \textbf{complete} iff it is admissible and all arguments defended by $E$ are in $E$.
\item \textbf{stable} iff it is conflict--free and for each $a\in A \setminus E$ there exists an argument $b \in E$ s.t. $(b,a) \in R$. 
\end{itemize}
The \textbf{characteristic function} $F_{AF} : 2^A \rightarrow 2^A$ is defined as:
$F_{AF}(E) = \{a \mid a \text{ is defended by } E \text{ in } AF\}$. The \textbf{grounded extension} is the least fixed point of $F_{AF}$.
\end{definition}

In the context of this paper, we would also like to recall the notion of range:
\begin{definition}
Let $E^+$ be the set of arguments attacked by $E$ and $E^-$ the set of arguments that attack $E$ . $E^+ \cup E$ is the \textbf{range} of $E$.
\end{definition}

Please note the concepts $E^+$ and the $E^-$ sets can be used to redefine defense. This idea
will be partially used in creating the semantics of ADFs.
Moreover, there is also an alternative way of computing the grounded extension: 
\begin{proposition}
The unique \textbf{grounded extension} of $AF$ is defined
as the outcome $E$ of the following “algorithm”. Let us start with $E=\emptyset$:
\begin{enumerate}
\item put each argument $a \in A$ which is not attacked in $AF$ into $E$; if no
such argument exists, return $E$.
\item remove from $AF$ all (new) arguments in $E$ and all arguments attacked
by them (together with all adjacent attacks) and continue with
Step 1.
\end{enumerate}
\label{prop:dung-grd}
\end{proposition}

What we have described above forms a family of the extension--based semantics. However, there exist also labeling--based ones \cite{CaminadaG09,article:semintro}.
 Instead of computing sets of accepted arguments, they generate
labelings, i.e. total functions $Lab: A \rightarrow \{in, out, undec\}$. Although we will not recall them here, we would like to draw the attention
to the fact that for every extension we can obtain an appropriate labeling and vice versa. This property is particularly important
as it does not fully carry over to the ADF setting.

Finally, we would like to recall several important lemmas and theorems from the original paper on AFs \cite{article:dung}. 

\begin{lemma}{\textbf{Dung's Fundamental Lemma}}
Let $E$ be an admissible extension, $a$ and $b$ two arguments defended by $E$. Then $E' = E \cup \{a\}$ is admissible and $b$ is defended by $E'$.
\end{lemma}

\begin{theorem}
Every stable extension is a preferred extension, but not vice versa.
Every preferred extension is a complete extension, but not vice versa.
 The grounded extension is the least w.r.t. set inclusion complete extension.
The complete extensions form a complete semilattice w.r.t. set inclusion. \footnote{A partial order $(A,\leq)$ is a complete semilattice iff each nonempty
subset of $A$ has a glb and each increasing sequence of $S$ has a lub.}

\end{theorem}
%
%
%

\section{Argumentation Frameworks with Support}
\label{sec:bip}

Currently the most recognized frameworks with support are the Bipolar Argumentation Framework BAF \cite{article:newbaf}, 
Argumentation Framework
with Necessities AFN \cite{incoll:newafn} and Evidential Argumentation System EAS \cite{inproc:eas}.
We will now briefly recall them in order to further motivate the directions of the semantics we have
taken in ADFs.

The original bipolar argumentation framework BAF \cite{incoll:bipolar} studied a relation we will refer to as abstract support:
\begin{definition}
A \textbf{bipolar argumentation framework} is a tuple $(A, R, S)$, where $A$ is a set of \textbf{arguments}, 
$R \subseteq A \times A$ represents the \textbf{attack} relation and $S \subseteq A \times A$ the \textbf{support}. 
\end{definition}

The biggest difference between this abstract relation and any other interpretation of support
 is the fact that it did not affect the acceptability of an argument, i.e. even a supported argument could be accepted "alone".
 The positive interaction was used to derive additional indirect forms of attack and based on them, stronger versions of conflict--freeness
were developed. 
\begin{definition}
We say that an argument $a$ \textbf{support attacks} argument $b$, if there exists some argument $c$
s.t. there is a sequence of supports from $a$ to $c$ (i.e. $a S...S c$) and $c R b$.
We say that $a$ \textbf{secondary attacks} $b$ if there is some argument $c$ s.t. $c S...S b$
and $a R c$.
We say that $B\subseteq A$ is:
\begin{itemize}
\item \textbf{+conflict--free} iff $\nexists a, b \in B$ s.t. $a$ (directly or indirectly) attacks $b$.
\item \textbf{safe} iff $\nexists b\in A$ s.t. $b$ is at the same time (directly or indirectly) attacked by $B$ and either there is a sequence of supports
from an element of $B$ to $b$, or $b \in B$. 
\item \textbf{closed under $S$} iff $\forall b\in B, a \in A$, if $bSa$ then $a\in B$.
\end{itemize}
\end{definition}

The definition of defense remains the same and any Dung semantics is specialized by choosing an given notion of conflict--freeness or safety. Apart
from the stable semantics, no assumptions as to cycles occurring in the support relation are made. The later developed deductive support \cite{inproc:support}
remains in the BAF setting and is also modeled by new indirect attacks \cite{article:newbaf}. Consequently, acyclicity is not required.
%
%

The most recent formulation of the framework with necessary support is as follows \cite{incoll:newafn}:

\begin{definition}
An \textbf{argumentation framework with necessities} is a tuple $(A, R, N)$, where $A$ is the set of \textbf{arguments}, $R \subseteq A \times A$ represents
(binary) \textbf{attacks},
and $N \subseteq (2^A \setminus \emptyset) \times A$ is the \textbf{necessity relation}. 
\end{definition}

Given a set $B\subseteq A$ and an argument $a$,
$B N a$ should be read as "at least one element of $B$ needs to be present in order to accept $a$". The AFN semantics are built around
the notions of coherence:

\begin{definition}
We say that a set of arguments $B$ is \textbf{coherent} iff every $b\in B$ is powerful, i.e. there exists a sequence $a_0,..,a_n$ of
some elements of $B$ s.t 
\begin{inparaenum}[\itshape 1\upshape)]
\item $a_n = b$, 
\item $\nexists C\subseteq A$ s.t. $CN a_0$, and 
\item for $1\leq i \leq n$ it holds that for every set $C\subseteq A$
if $CN a_i$, then $C \cap \{a_0,...,a_{i-1}\} \neq \emptyset$.
\end{inparaenum}
 A coherent set $B$ is \textbf{strongly coherent} iff it is conflict--free.
\end{definition}

Although it may look a bit complicated at first, the definition of coherence grasps the intuition that
we need to provide sufficient acyclic support for the arguments we want to accept. 
Defense in AFNs is understood as the ability to provide support and to counter the attacks from any coherent set.

\begin{definition} 
We say that a set
$B\subseteq A$ \textbf{defends} $a$, if $B\cup \{a\}$ is coherent and for every $c\in A$, if $c R a$ then for every coherent set $C\subseteq A$ containing $c$, 
$B R C$. 
\end{definition}

Using the notion of strong coherence and defense, the AFN semantics are built in a way corresponding to Dung semantics.
It is easy to see that, through the notion of coherency, AFNs discard cyclic arguments both on the "inside" and the "outside". This means we cannot 
accept them in an extension and they are not considered
as valid attackers.

The last type of support we will consider here is the the \textit{evidential support} \cite{inproc:eas}. It distinguishes 
between standard and \textit{prima facie} arguments. The latter are the only ones that are valid without any
support. Every other argument that we want to accept needs to be supported by at least one prima facie argument, be it directly or not.

\begin{definition}
An \textbf{evidential argumentation system} (EAS) is a tuple $(A, R, E)$ where $A$ is 
a set of \textbf{arguments}, $R \subseteq (2^A \setminus \emptyset) \times A$ is the \textbf{attack} relation,
and $E \subseteq (2^A \setminus \emptyset) \times A$ is the \textbf{support} relation. 
We distinguish a special argument
$\eta \in A$ s.t. $\nexists (x,y) \in R$ where $\eta \in x$; and $\nexists x$ where $(x, \eta) \in R$ or $(x, \eta) \in E$.
\end{definition}
$\eta$ represents the prima facie arguments and is referred to as evidence or environment.
The idea that the valid arguments (and attackers) need to trace back to it is captured with the notions of e--support and 
e--supported attack\footnote{The presented definition is slightly different from the one available in \cite{inproc:eas}. The new version was
obtained through personal communication with the author.}.

\begin{definition}
\label{def:esup}
An argument $a\in A$ has \textbf{evidential support} (e--support) from a set $S\subseteq A$ iff $a=\eta$ or
there is a non-empty $S' \subseteq S$ s.t. $S' E a$ and $\forall x \in S'$, $x$ has e--support from $S \setminus \{a\}$.
\end{definition}

\begin{definition}
\label{def:eatt}
A set $S\subseteq A$ carries out an \textbf{evidence supported attack} (e--supported attack)
 on $a$ iff $(S',a) \in R$ where $S' \subseteq S$, and for all $s \in S'$, $s$ has e--support from $S$. 
An e--supported attack by $S$ on $a$ is \textbf{minimal} iff there is no $S' \subset S$ that carries out an e--supported attack on $a$.
\end{definition}

The EASs semantics are built around the notion of acceptability in a manner similar to those of Dung's. However, in AFs only 
the attack relation was considered. In EASs, also sufficient support is required:

\begin{definition}
An argument $a$ is \textbf{acceptable} w.r.t. a set $S\subseteq A$ iff $a$ is e--supported by $S$ and
given a minimal e--supported attack by a set $T\subseteq A$ against $a$, it is the case that $S$ carries out an e--supported attack against a member of $T$.
\end{definition}

The notion of conflict--freeness is easily adapted to take set, not just binary conflict into account. With this and the notion of acceptability, the
EASs semantics are built just like AF semantics.
From the fact that every valid argument needs to be grounded in the environment it clearly results that EAS semantics are acyclic both on the inside and outside.

\section{Abstract Dialectical Frameworks}
\label{sec:adf} 

Abstract dialectical frameworks have been defined in \cite{inproc:adf} and further studied
 in \cite{tofix:newadf,inproc:adm,report:strass,strass13instantiating,strass-wallner14complexity}.
The main goal of ADFs is to be able to express arbitrary relations and avoid the need of extending AFs 
by new relation sets each time they are needed. This is achieved by the means of the acceptance conditions, which
 define what arguments should  
be present in order to accept or reject a given argument.
\begin{definition}
An \textbf{abstract dialectical framework} (ADF) as a tuple $(S, L, C)$, where
$S$ is a set of abstract \textbf{arguments} (nodes, statements),
$L \subseteq S \times S$ is a set of \textbf{links} (edges) and 
$C = \{C_ s \}_{s\in S}$ is a set of \textbf{acceptance conditions}, one condition per each argument. 
An acceptance condition is a total function $C_s : 2^{par(s)} \rightarrow \{in, out\}$, where
$par(s) = \{ p \in S \mid (p,s) \in L\}$ is the set of parents of an argument $s$.
\end{definition}

One can also represent the acceptance conditions by propositional formulas \cite{thesis:stefan} rather than functions. By this we mean
that given an argument $s \in S$, $C_s = \varphi_s$, where $\varphi_s$ is a propositional formula over arguments $S$.
As we will be making
use of both extension and labeling--based semantics, we need to provide necessary information on interpretations first (more details can be found in \cite{tofix:newadf,inproc:adm}).
Please note that the links in ADFs only represent connections between arguments, while the burden of deciding the nature of these
 connections falls to the acceptance conditions. Moreover,
parents of an argument can be easily extracted from the conditions. Thus, we will use of shortened notation $D = (S,C)$ through the rest of this paper.

\subsection{Interpretations and decisiveness}
\label{sec:premdec}

A two (or three--valued) interpretation is simply a mapping that assigns the truth values $\{\tvt, \tvf\}$ (respectively $\{\tvt, \tvf, \tvu\}$) to arguments. We will
be making use both of partial (i.e. defined only for a subset of $S$) and the full ones.
In the three--valued setting we will adopt the
precision (information) ordering of the values: $\tvu \leq_i \tvt$ and $\tvu \leq_i \tvf$
The pair
$(\{\textbf{t}, \textbf{f}, \textbf{u}\}, \leq_i)$ forms a complete meet--semilattice with the meet operation $\sqcap$ assigning values in the following way:
 $\textbf{t} \, \sqcap \, \textbf{t} = \textbf{t}$, $\textbf{f} \, \sqcap \, \textbf{f} = \textbf{f}$ and $\textbf{u}$ in all other cases.
It can naturally be extended to interpretations: given two interpretations $v$ and $v'$ on $S$, we say that $v'$ contains
more information, denoted $v \leq_i v'$, iff $\forall_{s\in S}\, v(s) \leq_i v'(s)$. Similar follows for the meet operation. In case $v$ is three and $v'$ two--valued, 
we say that $v'$ extends $v$.
This means that elements mapped originally to $\tvu$ are now assigned either $\tvt$ or $\tvf$. The
set of all two--valued interpretations extending $v$ is denoted $\lbrack v \rbrack_2$. 

\begin{example}
Let $v=\{a:\tvt, b:\tvt, c:\tvf, d:\tvu)$ be a three--valued interpretation. We have
two extending interpretations, $v'=\{a:\tvt, b:\tvt, c:\tvf, d:\tvt)$ and $v''=\{a:\tvt, b:\tvt, c:\tvf, d:\tvf)$.
Clearly, it holds that $v \leq_i v'$ and $v\leq_i v''$. However, $v'$ and $v''$ are incomparable w.r.t. $\leq_i$.

Let now $w=\{a:\tvf, b:\tvf, c:\tvf, d:\tvt)$ be another three--valued interpretation.  $v \sqcap w$
gives us a new interpretation $w' = \{a:\tvu, b:\tvu, c:\tvf, d:\tvu)$: as the assignments of $a,b$ and $d$ differ between $v$ and $w$,
the resulting value is $\tvu$. On the other hand, $c$ is in both cases $\tvf$ and thus retains its value.
\end{example}

We will use $v^x$ to denote a set of arguments mapped to $x$ by $v$, where $x$ is some truth--value.
Given an acceptance condition $C_s$ for some argument $s \in S$ and an interpretation $v$, 
we define a shorthand $v(C_s)$ as $C_s(v^\tvt \cap par(s))$. For a given propositional formula $\varphi$ and an interpretation $v$ defined over all of the atoms of the formula,
$v(\varphi)$ will just stand for the value of the formula under $v$. However, apart from knowing the "current" value of a given acceptance
condition for some interpretation, we would also like to know if this interpretation is "final". By this we understand that no new information will cause
the value to change.
This is expressed by the notion of decisive interpretations, which
are at the core of the extension--based ADF semantics.

\begin{definition}
Given an interpretation $v$ defined over a set $A$, \textbf{completion} of $v$ to a set $Z$ where $A\subseteq Z$ is
an interpretation $v'$ defined on $Z$ in a way that $\forall a \in A \; v(a) = v'(a)$. By a $\tvt/\tvf$ completion
we will understand $v'$ that maps all arguments in $Z\setminus A$ respectively to $\tvt/\tvf$.
\end{definition}

The similarity between the concepts of completion and extending interpretation should not be overlooked. Basically, given
a three--valued interpretation $v$ defined over $S$, the set $\lbrack v \rbrack_2$ precisely corresponds to the set of completions to $S$
of the two--valued part of $v$. However, the extension notion from the three--valued setting can be very misleading when used
in the extension--based semantics. Therefore, we would like to keep the notion of completion.

\begin{definition}
We say that a two--valued interpretation $v$ is \textbf{decisive} for
an argument $s \in S$ iff for any two completions
$v_{par(s)}$ and $v'_{par(s)}$ of $v$ to $A \cup par(s)$,  
it holds that $v_{par(s)}(C_s) =v'_{par(s)}(C_s)$.
We say that $s$ is \textbf{decisively out/in} wrt $v$ if $v$ is decisive and all of its completions evaluate $C_s$ to respectively $out,in$.
\end{definition}

\begin{example}
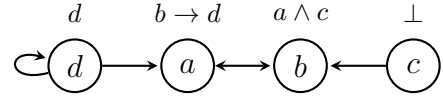
\begin{figure}[t]
\vspace{-1em}
\centering
\begin{tikzpicture}
[->,>=stealth,shorten >=1pt,auto,node distance=1.5cm,
  thick,main node/.style={circle,fill=none,draw,minimum size = 0.7cm,font=\Large\bfseries},
condition/.style={circle,fill=none,draw=none,minimum size = 0.3cm,font=\normalsize\bfseries}]

\node[main node] (a) {$a$};
\node[main node] (b) [right of=a] {$b$};
\node[main node] (c) [right of=b] {$c$};
\node[main node] (d) [left of =a] {$d$};
\node[condition](ca) [above of= a, yshift=-0.8cm] {$b \rightarrow d$};
\node[condition](cb) [above of= b, yshift=-0.8cm] {$a \land c$};
\node[condition](cc) [above of= c, yshift=-0.8cm] {$\bot$};
\node[condition](cd) [above of= d, yshift=-0.8cm] {$d$};

\draw [<->] (a.east)--(b.west){};
\draw [->] (c.west)--(b.east){};
\draw [->] (d.east)--(a.west){};
\path
	(d) edge [loop left] node {} (d);
\end{tikzpicture}
\caption{Sample ADF}
\label{fig:dec}\vspace{-1em}

\end{figure}

Let $(\{a,b,c,d\}, 
\{ \varphi_a: b\rightarrow d, \varphi_b:  a\land c, \varphi_c: \bot, \varphi_d:d\})$ be an ADF depicted in Figure \ref{fig:dec}.
Example of a decisively in interpretation for $a$ is $v=\{b: \tvf\}$. It simply means that knowing that $b$ is false, not matter the value of $d$,
the implication is always true and thus the acceptance condition satisfied. From the more technical side, it is the same as checking that both
completions to $\{b,d\}$, namely $\{b:\tvf, d:\tvt\}$  and $\{b:\tvf, d:\tvf\}$ satisfy the condition.
Example of a decisively out interpretation for $b$ is $v'=\{c: \tvf\}$. Again, it suffices to falsify one element of a conjunction to know that the whole formula
will evaluate to false. 
\end{example}

\subsection{Acyclicity}

Let us now focus on the issue of positive dependency cycles. Please note we refrain from calling them support cycles in the ADF setting
in order not to confuse them with specific definitions of support available in the literature \cite{article:newbaf}.

Informally speaking, an argument takes part in a cycle if its acceptance depends on itself. An intuitive way of verifying the acyclicity of an argument
would be to "track" its evaluation, e.g. in order to accept $a$ we need to accept $b$, to accept $b$ we need to accept $c$ and so on.
This basic case becomes more complicated when disjunction is introduced. We then receive a number of such "paths", with only some of them
proving to be acyclic. Moreover, they might be conflicting one with each other, and we can have a situation
in which all acyclic evaluations are blocked and a cycle is forced. Our approach to acyclicity
is based on the idea of such "paths" that are accompanied by sets of arguments used to detect possible conflicts.

Let us now introduce the formal definitions.
Given an argument $s\in S$ and $x\in\{in, out\}$, by $min\_dec(x,s)$ we will denote the set of minimal two--valued interpretations
that are decisively $x$ for $s$.  By minimal we understand that both $v^{\tvt}$ and  $v^{\tvf}$ are minimal w.r.t. set inclusion.

\begin{definition}
 Let $A \subseteq S$ be a nonempty set of arguments. 
A \textbf{positive dependency function} on $A$ is a function $pd$ assigning every argument 
$a \in A$ an interpretation $v \in min\_dec(in, a)$ s.t. $v^t \subseteq A$ or $\mathcal{N}$ (null) iff no such
interpretation can be found.
\end{definition}
\begin{definition}
An \textbf{acyclic positive dependency evaluation} $ace^a$ for $a \in A$ based on a given pd--function $pd$ is a pair $((a_0,...,a_n), B)$, 
\footnote{Please note that it is not required that $B\subseteq A$}
where $B = \bigcup_{i=0}^n \, pd(a_i)^{\tvf}$ and $(a_0,...,a_n)$ is
a sequence of distinct elements of $A$ s.t.:
\begin{inparaenum}[\itshape 1\upshape)]
\item $\forall_{i=0}^n \, pd(a_i) \neq \mathcal{N}$,
\item $a_n = a$,
\item $pd(a_0)^\tvt = \emptyset$, and
\item $\forall_{i=1}^n, \, pd(a_i)^{\tvt} \subseteq \{a_0,...,a_{i-1}\}$.
\end{inparaenum}
We will refer to the sequence part of the evaluation as \textbf{pd--sequence} and to the $B$ as the \textbf{blocking set}.
We will say that an argument $a$ is \textbf{pd--acyclic} in $A$ iff there exist a pd--function on $A$ and a corresponding acyclic pd--evaluation for $a$.
\end{definition}

We will  write that an argument has an acyclic pd--evaluation on $A$ if there is some pd--function on $A$ from which we can produce
the evaluation. There are two ways we can "attack" an acyclic evaluation. We can either discard an argument required by the evaluation
or accept one that is capable of preventing it. This corresponds to rejecting a member of a pd--sequence or accepting an argument
from the blocking set. We can now formulate this "conflict" by the means of an interpretation:

\begin{definition}
Let $A\subseteq S$ be a set of arguments and $a \in A$ s.t. $a$ has an acyclic pd--evaluation $ace^a = ((a_0,...,a_n), B) $ in $A$. We say that a
two--valued interpretation $v$ \textbf{blocks} $ace^a$ iff $\exists b\in B$ s.t. $v(b) = \tvt$ or $\exists a_i \in  \{a_0,...,a_n\}$ s.t. $v(a_i) = \tvf$.
\end{definition}

Let us now show on an example why we require minimality on the chosen interpretations and why do we store the blocking set:

\begin{example}[label=example1]
Let us assume an ADF $(\{a,b,c\}, \{C_a:\neg c \lor b, C_b:a, C_c:c\})$ depicted in Figure \ref{fig:cf}.
For argument $a$ there exist the following decisively in interpretations: $v_1 = \{c:\tvf\},v_2 = \{b:\tvt\},v_3 = \{b:\tvt, c:\tvf\},v_4= \{b:\tvt, c:\tvt\},
v_5 = \{b:\tvf, c:\tvf\}$. Only the first two are minimal. Considering $v_4$ would give us a wrong view that $a$ requires $c$ for acceptance, which is not
a desirable reading. The interpretations for $b$ and $c$ are respectively $w_1 = \{a:\tvt\}$ and $z_1 = \{c:\tvt\}$.
Consequently, we have two pd--functions on $\{a,b,c\}$, namely $pd_1 = \{a:v_1, b:w_1, c:z_1\}$ and $pd_2 = \{a:v_2, b:w_1, c:z_1\}$.
From them we obtain one acyclic pd--evaluation for $a$: $((a),\{c\})$, one for $b$: $((a,b),\{c\})$ and none for $c$.

Let us look closer at a set $E=\{a,b,c\}$. We can see that $c$ is not pd--acyclic in $E$. However, the presence of $c$ also "forces" a cycle between $a$ and $b$. 
The acceptance conditions of all arguments are satisfied, thus this simple check is not good enough to verify if a cycle occurs. 
Only looking at the whole evaluations
shows us that $a$ and $b$ are both blocked by $c$. Although $a$ and $b$ are pd--acyclic in $E$,
we see that their evaluations are in fact blocked and this second level of conflict needs to be taken into account by the semantics.
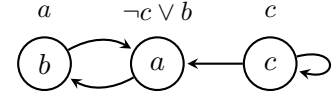
\begin{figure}[t]
\vspace{-1em}
\centering
  \begin{tikzpicture}
[->,>=stealth,shorten >=1pt,auto,node distance=1.5cm,
  thick,main node/.style={circle,fill=none,draw,minimum size = 0.7cm,font=\large\bfseries},
condition/.style={circle,fill=none,draw=none,minimum size = 0.3cm,font=\normalsize\bfseries}]

\node[main node] (a) {$a$};
\node[main node] (b) [left of=a] {$b$};
\node[main node] (c) [right of=a] {$c$};

\node[condition](ca) [above of= a, yshift=-0.8cm] {$\neg c \lor b$};
\node[condition](cb) [above of= b, yshift=-0.8cm] {$a$};
\node[condition](cc) [above of= c, yshift=-0.8cm] {$c$};

 \path
	(a) edge  [bend left] node{} (b)
	(b) edge  [bend left] node{} (a)
	(c) edge  node{} (a)	
(c) edge [loop right] node{} (c);
\end{tikzpicture}
\caption{Sample ADF}
\label{fig:cf}\vspace{-1em}
\end{figure}
\label{ex:cf}
\end{example}

As a final remark, please note that it can be the case that an evaluation is self--blocking.
We can now proceed to recall existing and introduce new semantics of the abstract dialectical frameworks.

\section{Extension--Based Semantics of ADFs}
\label{sec:sem}

Although various semantics for ADFs have already been defined in the original paper \cite{inproc:adf}, only three of them --
conflict--free, model and grounded (initially
referred to as well--founded) -- are still used (issues with the other formulations can be found in
\cite{tofix:newadf,inproc:adm,report:strass}). Moreover, the treatment of cycles and their handling by the semantics was not sufficiently developed.
In this section we will address all of those issues. Before we continue, let us first motivate our choice on how to treat cycles.
The opinions on support cycles differ between the available frameworks, as we have shown in Section \ref{sec:bip}.
Therefore, we would like to explore the possible approaches in the context of ADFs by developing appropriate semantics. 

The classification of the sub--semantics that we will adopt in this paper is based on the inside--outside intuition we presented
in the introduction. Appropriate semantics will receive a two--element prefix $xy-$, where $x$ will denote whether cycles
are permitted or not on the "inside" and $y$ on the "outside". We will use $x, y \in \{a,c\}$, where $a$ will stand for \textit{acyclic} and $c$ for \textit{cyclic}
constraints.
%
As the conflict--free (and naive) semantics focus
only on what we can accept, we will drop the prefixing in this case. Although the model, stable and grounded fit into our classification (more
details can be found in this section and in \cite{report:semantics}), they have a sufficiently unique naming and further annotations are not
necessary. We are thus left with admissible, preferred and complete. The BAF approach follows the idea that we can accept
arguments that are not acyclic in our opinion and we allow our opponent to do the same. The ADF semantics we have developed in 
\cite{inproc:adm} also shares this view. Therefore, they will receive the $cc-$ prefix. On the other hand, AFN and EAS semantics
do not permit cycles both in extensions and as attackers. Consequently, the semantics following this line of reasoning
will be prefixed with $aa-$.
Please note we believe that a non--uniform approach can also be suitable in certain situations. By non--uniform we mean not accepting cyclic arguments, but still treating
them as valid attackers and so on (i.e. $ca-$ and $ac-$). However, in this paper we would like to focus only on the two perspectives mentioned before.

\subsection{Conflict--free and naive semantics}

In the Dung setting, conflict--freeness meant that the elements of an extension could not attack one another. 
 Providing an argument with the required support is then a separate condition in frameworks such as AFNs and EASs. In ADFs, where we lose the set representation
of relations
in favor of abstraction, not including "attackers" and accepting "supporters" is combined into one notion. This represents the intuition of arguments that can stand
together presented in \cite{article:semintro}. Let us now assume an ADF $D= (S,C)$.
\begin{definition}
A set of arguments $E \subseteq S$ is \textbf{conflict--free} in $D$ iff for all 
$s \in E$ we have $C_s (E \cap par(s )) = in$.
\end{definition}

In the acyclic version of conflict--freeness we also need to deal with the conflicts arising on the level of evaluations.
 To meet the formal requirements, we first have to show how the notions of range and the $E^+$ set are moved to ADFs.
\begin{definition}
Let $E\subseteq S$ a conflict--free extension of $D$ and $v_E$ a partial two--valued interpretation built  as follows:
\begin{enumerate}
\item Let $M=E$ and for every $a\in M$ set $v_E(a)=\tvt$; 
\item For every argument $b \in S\setminus M$ that is decisively out in $v_E$, set $v_E(b)=\tvf$ and add $b$ to $M$;
\item Repeat the previous step until there are no new elements added to $M$. 
\end{enumerate}
By $E^+$ we understand the set of arguments $v_E^\tvf$ and we will refer to it as the \textbf{discarded set}. $v_E$ now forms the \textbf{range interpretation} of $E$.
\end{definition}

However, the notions of the discarded set and the range are quite strict in the sense that they require an explicit "attack" on arguments that take part in
dependency cycles. This is not always a desirable property. Depending on the approach we might not treat cyclic arguments as valid and hence
want them "out of the way". 

\begin{definition}
Let $E\subseteq S$ a conflict--free extension of $D$ and $v_E^a$ a partial two--valued interpretation built  as follows:
\begin{enumerate}
\item Let $M=E$. For every $a\in M$ set $v_E^a(a)=\tvt$.
\item For every argument $b \in S\setminus M$ s.t.
every acyclic pd--evaluation of $b$ in $S$ is blocked by $v_E^a$, set $v_E^a(b)=\tvf$ and add $b$ to $M$.
\item Repeat the previous step until there are no new elements added to $M$. 
\end{enumerate}
By $E^{a+}$ we understand the set of arguments mapped to $\tvf$ by $v_E^a$ and refer to it as \textbf{acyclic discarded set}.
We refer to $v_E^a$ as \textbf{acyclic range interpretation} of $E$.
\end{definition}
%
%

We can now define an acyclic version of conflict--freeness:
\begin{definition}
A conflict--free extension $E$ is a
 \textbf{pd--acyclic conflict--free extension} of $D$ iff every
argument $a \in E$ has an unblocked acyclic pd--evaluation on $E$ w.r.t. $v^E$.
\end{definition}

As we are dealing with a conflict-- free extension, all the arguments of a given pd--sequence are naturally $\tvt$ both in $v_E$ and $v_E^a$.
Therefore, in order to ensure that an evaluation $((a_0,...,a_n), B)$ is unblocked it suffices to check whether $E \cap B = \emptyset$. Consequently, 
in this case it does not matter w.r.t. to which version of range we are verifying the evaluations.

\begin{definition}
The \textbf{naive}
and \textbf{pd--acyclic naive} extensions are respectively maximal w.r.t. set inclusion conflict--free and pd--acyclic conflict--free extensions.
\end{definition}

\begin{example}[continues=example1]
Recall the ADF $(\{a,b,c\}, \{C_a:\neg c \lor b, C_b:a, C_c:c\})$.
The conflict--free extensions are $\emptyset, \{a\}, \{c\}, \{a,b\}$ and $\{a,b,c\}$. Their standard discarded set in all cases is just $\emptyset$ -- none
of the sets has the power to decisively out the non--members. The acyclic discarded set of $\emptyset$, $\{a\}$ and $\{a,b\}$ 
is now $\{c\}$, since it has no acyclic evaluation to start with. In the case of $\{c\}$, it is $\{a,b\}$, which is to be expected since $c$ had the power
to block their evaluations. Finally, $\{a,b,c\}^{a+}$ is $\emptyset$. 
In the end, only $\emptyset, \{a\}$ and $\{a,b\}$ qualify for acyclic type.
The naive and pd--acyclic naive extensions are respectively $\{a,b,c\}$ and $\{a,b\}$.
\end{example}

\subsection{Model and stable semantics} 

The concept of a model basically follows the intuition that if something can be accepted, 
it should be accepted:
\begin{definition}
 A conflict--free extension $E$ is a \textbf{model} 
of $D$ if $\forall \;s \in S, \; \;C_s (E \cap par (s )) = in$ implies $s \in E$.
\end{definition}

 Although the semantics is simple, several of its properties should be explained. First of all, given a
model candidate $E$, checking whether a condition of some argument $s$ is satisfied does not verify if an argument depends on itself or if it "outs" a previously
included member of $E$. This means that an argument we should include may break conflict--freeness of the set. On the other hand, an argument
can be $out$ due to positive dependency cycles, i.e. its supporter is not present. And since model makes no acyclicity assumptions on the inside, 
arguments outed this way can later appear in a model $E\subset E'$. Consequently, it is clear to see that model 
semantics is not universally defined and the produced extensions might not be maximal w.r.t. subset inclusion.

The model semantics was used as a mean to obtain the stable models. The main idea was to make sure that the model is acyclic. Unfortunately, the used
reduction method was not adequate, as shown in \cite{tofix:newadf}. However, the initial idea still holds and we use it to define stability. Although the 
produced extensions are now incomparable w.r.t. set inclusion, the semantics is still not universally defined.
\begin{definition}
A model $E$ is a \textbf{stable extension} iff it is pd--acyclic conflict--free. 
\end{definition}
\begin{example}[continues=example1]
Let us again come back to the ADF $(\{a,b,c\}, \{C_a:\neg c \lor b, C_b:a, C_c:c\})$.
The conflict--free extensions were $\emptyset, \{a\}, \{c\}, \{a,b\}$ and $\{a,b,c\}$. The first two are not models, as in the first case
$a$ and in the latter $b$ can be accepted. Recall that  $\emptyset, \{a\}$ and $\{a,b\}$
were the pd--acyclic conflict--free extensions. The only one that is also a model is $\{a,b\}$ and thus we obtain our single stable extension.
\label{ex:mod}
\end{example}

\subsection{Grounded semantics}

Next comes the grounded semantics \cite{inproc:adf}. 
Just like in the Dung setting, it preserves the unique--status property, i.e. produces only a single extension. Moreover, it is
defined in the terms of a special operator:

\begin{definition}
 Let
$\Gamma'_D(A,R) = (acc(A, R ),reb (A, R ))$, where
 $acc(A, R)=\{ r\in S \mid A \subseteq S' \subseteq (S\backslash R) \Rightarrow
C_r(S' \cap par(r))=in\}$
and
$reb(A, R )=\{ r\in S \mid A \subseteq S' \subseteq (S\backslash R) \Rightarrow
C_r(S' \cap par(r))=out\}$. Then
$E$ is the \textbf{grounded model} of $D$ iff for some $E' \subseteq S, (E,E')$ is the least fix--point of $\Gamma'_D$. 
\end{definition}

Although it might look complicated at first, this is nothing more than analyzing decisiveness using a set, not interpretation form (please
see \cite{report:semantics} for more details). Thus, one can also obtain the grounded extension by an ADF version of
Proposition \ref{prop:dung-grd}:

\begin{proposition}
Let $v$ be an empty interpretation. For every argument $a\in S$ that is decisively in w.r.t. $v$, set $v(a) = \tvt$ and for every argument
$b\in S$ that is decisively w.r.t. $v$, set $v(b) = \tvf$. Repeat the procedure until no further assignments can be done. The \textbf{grounded extension}
of $D$ is then $v^\tvt$.
\label{prop:adf-grd}
\end{proposition}

\begin{example}[continues=example1]
Recall our ADF $(\{a,b,c\}, \{C_a:\neg c \lor b, C_b:a, C_c:c\})$.
 Let $v$ be an empty interpretation. It is easy to see that no argument is decisively in/out w.r.t. $v$. If we analyze $a$, 
it is easy to see that if we accept $c$, the condition is out, but if we accept both $b$ and $c$ it is in again. 
Although both $b$ and $c$ are out in $v$, the condition of $b$ can be met if we
accept $a$, and condition of $c$ if we accept $c$. Hence, we obtain no decisiveness again.
Thus, $\emptyset$ is the grounded extension.
\end{example}

\subsection{Admissible and preferred semantics}

In \cite{inproc:adm} we have presented our first definition of admissibility, before the sub--semantics classification was developed.
The new, simplified version of our previous formulation, is now as follows:

\begin{definition}
A conflict--extension $E \subseteq S$ is \textbf{cc--admissible} in $D$ iff every element of $E$ is decisively in w.r.t to its range
interpretation $v_E$. 
\end{definition}

It is important to understand how decisiveness encapsulates the defense known from the Dung setting. If an argument is decisively in, then any set of arguments
that would have the power to out the acceptance condition is "prevented" by the interpretation. Hence, the statements required for the acceptance of $a$
are mapped to $\tvt$ and those that would make us reject $a$ are mapped to $\tvf$. The former encapsulates the required support, while the latter 
contains the "attackers" known from the Dung setting.

When working with the semantics
that have to be acyclic on the "inside", we not only have to defend the members, but also their acyclic evaluations:

\begin{definition}
A pd--acyclic conflict--free extension $E$ is \textbf{aa--admissible} iff every argument in $E$ \begin{inparaenum}[\itshape 1\upshape)]
\item is decisively in w.r.t. acyclic range interpretation $v_E^a$, and
\item has an unblocked acyclic pd--evaluation on $E$ s.t. all members of its blocking set $B$ are mapped to $\tvf$ by $v_E^a$.
\end{inparaenum}
\label{def:adm1}
\end{definition}

\begin{definition}
A set of arguments is \textbf{xy--preferred} iff it is maximal w.r.t. set inclusion xy--admissible.
\end{definition}

The following example shows that decisiveness encapsulates defense of an argument, but not necessarily of its evaluation:

\begin{example}
Let us modify the ADF depicted in Figure \ref{fig:cf} by changing the condition of $c$: $(\{a,b,c\}, \{C_a:\neg c \lor b, C_b:a, C_c:\top\})$.
The new pd--evaluations are $((a),\{c\})$ for $a$, $((a,b),\{c\})$ for $b$ and $((c), \emptyset)$ for $c$.
The conflict--free extensions are now $\emptyset, \{a\}, \{c\}, \{a, b\}$ and $\{a, b, c\}$.
Apart from the last, all are pd--acyclic conflict--free. $\emptyset$ and $\{c\}$
are trivially both aa and cc--admissible and $\{a,b,c\}$ cc--admissible. The standard and acyclic discarded sets of $\{a\}$ are both
empty, thus $a$ is not decisively in (we can always utter $c$) and the set is neither aa
nor cc--admissible. The discarded sets of $\{a,b\}$ are also empty; however, it is easy to see
that both $a$ and $b$ are decisively in. Although uttering $c$ would not change the values of acceptance
conditions, it blocks the pd--evaluations of $a$ and $b$. Thus, $\{a,b\}$ is cc, but not aa--admissible.
The cc and aa--preferred extensions are respectively $\{a,b,c\}$ and $\{c\}$.
\end{example}

\begin{example}[continues=example1]
Let us come back to the original ADF $(\{a,b,c\}, \{C_a:\neg c \lor b, C_b:a, C_c:c\})$.
$\emptyset, \{a\}, \{c\}, \{a,b\}$ and $\{a,b,c\}$ were the standard and $\emptyset, \{a\}, \{a,b\}$ pd--acyclic conflict--free extensions.
$\emptyset$ is trivially both aa and cc, while $\{c\}$ and $\{a,b,c\}$ cc--admissible. The standard discarded sets of $\{a\}$ and $\{a,b\}$ are both empty,
while the acyclic ones are $\{c\}$. Consequently, $\{a\}$ is aa, but not cc--admissible. $\{a,b\}$ is both, but for different reasons;
in the cc--case, all arguments are decisively in (due to cyclic defense). In aa--approach, they are again decisively in, but the evaluations
are "safe" only because $c$ is not considered a valid attacker.
\label{ex:adm}
\end{example}

\subsection{Complete semantics}

Completeness represents an approach in which we have to accept everything we can safely conclude from our opinions. 
In the Dung setting, "safely" means defense, while in the bipolar setting it is strengthened by providing sufficient support.
In a sense, it follows the model intuition that what we can accept, we should accept. However, now we not only
use an admissible base in place of a conflict--free one, but also defend the arguments in question.
 Therefore, instead of checking if
 an argument is in, we want it to be decisively in.

\begin{definition}
A cc--admissible extension $E$ is \textbf{cc--complete} in $D$ iff every argument in $S$ that is decisively in w.r.t. to 
range interpretation $v_E$ is in $E$. 
\end{definition}

\begin{definition}
An aa--admissible extension $E$ is \textbf{aa--complete} in $D$ iff every argument in $S$ that is decisively in w.r.t. to acyclic
range interpretation $v_E^a$
is in $E$.
\end{definition}

Please note that in the case of aa--complete semantics, no further "defense" of the evaluation is needed, as visible in 
AA Fundamental Lemma (i.e. Lemma \ref{fund4}). This comes
from the fact that if we already have a properly "protected" evaluation, then appending a decisively in argument to it is sufficient for creating
an evaluation for this argument.

\begin{example}[continues=example1]

Let us now finish with the ADF $(\{a,b,c\}, \{C_a:\neg c \lor b, C_b:a, C_c:c\})$.
It is easy to see that all cc--admissible extensions are also cc--complete. However, only $\{a,b\}$ is aa--complete. Due to the fact
that $c$ is trivially included in any discarded set, $a$ can always be accepted (thus, $\emptyset$ is disqualified). Then, from acceptance of $a$,
acceptance of $b$ follows easily and $\{a\}$ is disqualified.
\end{example}

%
%
%
%
%

\subsection{Properties and examples}

%
Although the study provided here will by not be exhaustive,
we would like to show how the lemmas and theorems from the original paper on AFs \cite{article:dung} are shifted into this new setting.
The proofs can be found in \cite{report:semantics}.

Even though every pd--acyclic conflict--free extension is also conflict--free, it does not mean that every aa--admissible is cc--admissible.
These approaches differ significantly. The first one makes additional restrictions on the "inside", but due to acyclicity
requirements on the "outside" there are less arguments a given extension has to defend from. The latter allows more freedom as to what we can
accept, but also gives this freedom to the opponent, thus there are more possible attackers. Moreover, it should not come as a surprise
that these differences pass over to the preferred and complete semantics, as visible in Example \ref{ex1}.
Our results show that admissible sub--semantics satisfy the Fundamental Lemma. 

\begin{lemma}{CC Fundamental Lemma:}
Let $E$ be a cc--admissible extension, $v_E$ its range interpretation and $a, b\in S$ two arguments decisively in w.r.t. $v_E$. Then $E' = E \cup \{a\}$ is cc--admissible
and $b$ is decisively in w.r.t. $v_E'$.
\label{fund1}
\end{lemma}

\begin{lemma}{AA Fundamental Lemma:}
Let $E$ be an aa-admissible extension, $v_E^a$ its acyclic range interpretation and $a, b\in S$ two arguments decisively in w.r.t. $v_E^a$. 
Then $E' = E \cup \{a\}$ is aa--admissible and $b$ is decisively in w.r.t. $v_E'$.
\label{fund4}
\end{lemma}

The relations between the semantics presented in \cite{article:dung} are preserved by some of the specializations:

\begin{theorem}
 Every stable extension is an aa--preferred extension, but not vice versa.
 Every xy--preferred extension is an xy--complete extension for $x,y \in \{a,c\}$, but not vice versa.
The grounded extension might not be an aa--complete extension.
The grounded extension is the least w.r.t. set inclusion cc--complete extension.
\end{theorem}

\begin{example}[label=ex1]
Let $(\{a,b,c,d\},\{C_a: \neg b, C_b: \neg a, C_c: b\land \neg d, C_d:d \})$ be the ADF depicted in Figure \ref{fig:adf1}. The obtained
extensions are visible in Table \ref{tab:ext}. The conflict--free, model, stable, grounded, admissible, complete and preferred semantics
will be abbreviated to CF, MOD, STB, GRD, ADM, COMP and PREF.  The prefixing is visible in second column. 
In case of conflict--freeness, $C$ will denote the standard, and $A$
the pd--acyclic one.
\begin{figure}[t]
\vspace{-1em}
\centering
  \begin{tikzpicture}
[->,>=stealth,shorten >=1pt,auto,node distance=1.5cm,
  thick,main node/.style={circle,fill=none,draw,minimum size = 0.7cm,font=\large\bfseries},
condition/.style={circle,fill=none,draw=none,minimum size = 0.3cm,font=\bfseries}]

\node[main node] (a) {$a$};
\node[main node] (b) [right of=a] {$b$};
\node[main node] (c) [right of=b] {$c$};
\node[main node] (d) [right of=c] {$d$};

\node[condition](ca) [above of= a, yshift=-0.8cm] {$\neg b$};
\node[condition](cb) [above of= b, yshift=-0.8cm] {$\neg a$};
\node[condition](cc) [above of= c, yshift=-0.8cm] {$b \land \neg d$};
\node[condition](cd) [above of= d, yshift=-0.8cm] {$d$};

 \path
	(b) edge node {} (c)
	(b) edge  [bend left] node{} (a)
	(a) edge  [bend left] node{} (b)
	(d) edge node{} (c)
	(d) edge [loop right] node{} (d);
\end{tikzpicture}
\caption{Sample ADF}
\label{fig:adf1}
\end{figure}
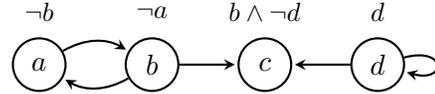
\begin{table}[h]
\caption{Extensions of the ADF from Figure \ref{fig:adf1}.}

\centering
\begin{tabular}{|c ||c|l| }
\hline
\multirow{2}{*}{CF}  
& $C$ & $\emptyset$, $\{a\}$, $\{b\}$, $\{d\}$, $\{b,c\}$, $\{a,d\}$, $\{b,d\}$ \\
& $A$ & $\emptyset$, $\{a\}$, $\{b\}$, \st{$\{d\}$}, $\{b,c\}$, \st{$\{a,d\}$, $\{b,d\}$} \\
\hline
MOD && \st{$\emptyset$}, $\{a\}$, \st{$\{b\}$, $\{d\}$}, $\{b,c\}$, $\{a,d\}$, $\{b,d\}$\\
\hline
STB && \st{$\emptyset$}, $\{a\}$, \st{$\{b\}$, $\{d\}$}, $\{b,c\}$, \st{$\{a,d\}$, $\{b,d\}$} \\
\hline
GRD &&  $\emptyset$, \st{$\{a\}$, $\{b\}$, $\{d\}$, $\{b,c\}$, $\{a,d\}$, $\{b,d\}$} \\
\hline
\multirow{2}{*}{ADM} 
 & $CC$ & $\emptyset$, $\{a\}$, $\{b\}$, $\{d\}$, \st{$\{b,c\}$}, $\{a,d\}$, $\{b,d\}$ \\
 & $AA$ & $\emptyset$, $\{a\}$, $\{b\}$, \st{$\{d\}$}, $\{b,c\}$, \st{$\{a,d\}$, $\{b,d\}$} \\ 
\hline
\multirow{2}{*}{COMP} 
 & $CC$ & $\emptyset$, $\{a\}$, $\{b\}$, $\{d\}$, \st{$\{b,c\}$}, $\{a,d\}$, $\{b,d\}$ \\
 & $AA$ & $\emptyset$, $\{a\}$, \st{$\{b\}$, $\{d\}$}, $\{b,c\}$, \st{$\{a,d\}$, $\{b,d\}$} \\ 
\hline
\multirow{2}{*}{PREF} 
 & $CC$ & \st{$\emptyset$, $\{a\}$, $\{b\}$, $\{d\}$, $\{b,c\}$}, $\{a,d\}$, $\{b,d\}$ \\
 & $AA$ & \st{$\emptyset$}, $\{a\}$, \st{$\{b\}$, $\{d\}$}, $\{b,c\}$, \st{$\{a,d\}$, $\{b,d\}$} \\ 
\hline
\end{tabular}
\label{tab:ext}
\end{table}

\end{example}

\section{Labeling--Based Semantics of ADFs}
\label{sec:comparison}

The two approaches towards labeling--based semantics of ADFs were developed in \cite{report:strass,tofix:newadf}. 
We will focus on the latter one, based on the notion of a three--valued characteristic operator:
\begin{definition}
Let $V_S$ be the set of all three--valued interpretations defined on $S$, $s$ and argument in $S$ and $v$ an interpretation in $V_S$.
The \textbf{three--valued characteristic operator} of $D$ is a function $\Gamma_D: V_S \rightarrow V_S$ s.t.
$\Gamma_D(v) = v'$ with 
$v'(s) = \bigsqcap_{w \in \lbrack v \rbrack_2} C_s(par(s) \cap w^\tvt)$.
\end{definition}

Verifying the value of an acceptance condition under a set of extensions$\lbrack v \rbrack_2$ of a three--valued interpretation $v$
 is exactly checking its value in the completions
of the two--valued part of $v$. Thus, an argument that is $\tvt$/$\tvf$ in $\Gamma_D(v)$ is decisively in/out w.r.t. to the two--valued part of $v$.

It is easy to see that in a certain sense this operator allows self--justification and self--falsification, i.e. that status of an argument depends on itself.
 Take, for example, a self--supporter;
if we generate an interpretation in which it is false then, obviously, it will remain false. Same follows if we assume it to be true.
This results from the fact that the operator functions on interpretations defined on all arguments, thus allowing
a self--dependent argument to affect its status.

The labeling--based semantics are now as follows:
\begin{definition}
 Let $v$ be a three--valued interpretation for $D$ and $\Gamma_D$ its characteristic operator. We say that $v$ is:
\begin{itemize}
\item \textbf{three--valued model} iff for all $s \in S$ we have that $v(s) \neq \tvu$ implies that $v(s) = v(\varphi_s)$; 
\item \textbf{admissible} iff $v \leq_i \Gamma_D(v)$; 
\item \textbf{complete} iff $v = \Gamma_D(v)$; 
\item \textbf{preferred} iff it is $\leq_i$--maximal admissible;
\item \textbf{grounded} iff it is the least fixpoint of $\Gamma_D$.
\end{itemize}
\end{definition}
Although in the case of stable semantics we formally
receive a set, not an interpretation, this difference is not significant. As nothing is left undecided, we can safely map all remaining arguments to
$\tvf$.
The current state of the art definition \cite{report:strass,tofix:newadf} is as follows:

\begin{definition}
Let $M$ be a model of $D$. A \textbf{reduct} of $D$ w.r.t. $M$ is $D^M = (M, L^M, C^M)$, where $L^M = L \cap (M \times M)$ and for
$m \in M$ we set $C_m^M = \varphi_m\lbrack b/\tvf: b \notin M\rbrack$. Let $gv$ be the grounded model of $D^M$.
Model $M$ is \textbf{stable} iff $M = gv^{\tvt}$. 
\end{definition}

\begin{example}[continues=ex1]
Let us now compute the possible labelings of our ADF. As there are over
twenty possible three--valued models, we will not list them. We have in total 15 admissible interpretations:
$v_1 = \{a: \tvf,b: \tvt,c: \tvu,d: \tvt\}, 
v_2 = \{a: \tvt,b: \tvf,c: \tvu,d: \tvu\},
v_3 = \{a: \tvu,b: \tvu,c: \tvu,d: \tvt\}, 
v_4 = \{a: \tvt,b: \tvf,c: \tvu,d: \tvt\}, 
v_5 = \{a: \tvf,b: \tvt,c: \tvu,d: \tvf\},
v_6 = \{a: \tvt,b: \tvf,c: \tvu,d: \tvf\},
v_7 = \{a: \tvu,b: \tvu,c: \tvu,d: \tvu\},
v_8 =  \{a: \tvu,b: \tvu,c: \tvf,d: \tvt\},
v_9 = \{a: \tvt,b: \tvf,c: \tvf,d: \tvt\}, 
v_{10} = \{a: \tvf,b: \tvt,c: \tvt,d: \tvf\}, 
v_{11}= \{a: \tvu,b: \tvu,c: \tvu,d: \tvf\}, 
v_{12} =  \{a: \tvt,b: \tvf,c: \tvf,d: \tvu\}, 
v_{13} = \{a: \tvf,b: \tvt,c: \tvu,d: \tvu\},
v_{14} = \{a: \tvf,b: \tvt,c: \tvf,d: \tvt\}$ and 
$v_{15} =  \{a: \tvt,b: \tvf,c: \tvf,d: \tvf\}$.
Out of them $v_7$ to $v_{15}$ are complete. The ones
that maximize the information content in this case are the ones without any $\tvu$
mappings: $v_9$, $v_{10}$, $v_{14}$ and $v_9$. $v_{10}$ and $v_{15}$ are stable
and finally, $v_7$ is grounded.
\label{ex1lab}
\end{example}

\subsection{Comparison with the extension--based approach}

We will start the comparison of extensions and labelings 
by relating conflict--freeness and three--valued models. Please note that the intuitions of two--valued and three--valued models
are completely different and should not be confused.
We will say that an extension $E$ and a labeling
$v$ correspond iff $v^\tvt = E$. 

\begin{theorem}
Let $E$ be a conflict--free and $A$ a pd--acyclic conflict--free extension. The $\tvu$--completions of $v_E$, $v_A$ and $v_A^a$
are three--valued models.
\end{theorem}

Let us continue with the admissible semantics. First, we will tie the notion of decisiveness to admissibility, following the comparison of completions and extending
interpretations that we have presented in Section \ref{sec:premdec}. 

\begin{theorem}
Let $v$ be a three--valued interpretation and $v'$ its (maximal) two--valued sub--interpretation. $v$ is admissible iff all arguments mapped to $\tvt$ are
decisively in w.r.t. $v'$ and all arguments mapped to $\tvf$ are decisively out w.r.t. $v'$.
\label{thm:declab}
\end{theorem}

Please note that this result does not imply that admissible extensions and labelings "perfectly" coincide.
 In labelings, we guess an interpretation, and thus assign initial values to arguments
that we want to verify later. If they are self--dependent, it of course affects the outcome. 
 In the extension based approaches, we distinguish whether
this dependency is permitted. Therefore, the aa-- and cc-- approaches will
have a corresponding labeling, but not vice versa. 

\begin{theorem}
Let $E$ be a cc--admissible and $A$ an aa--admissible extension. The $\tvu$--completions of $v_E$ and $v_A^a$ are admissible labelings.
\label{thm:extlabadm}
\end{theorem}

Let us now consider the preferred semantics. Information maximality is not the same as maximizing the set of accepted arguments and
due to the behavior of $\Gamma_D$ we can obtain a preferred interpretation that can map to $\tvt$ a subset of arguments of
another interpretation.
Consequently, we fail to receive an exact correspondence between the semantics. By this we mean
that given a framework there can exist an (arbitrary) preferred extension without a labeling counterpart and a labeling without an appropriate
extension of a given type.

\begin{theorem}
For any xy--preferred extension there might not exist a corresponding preferred labeling and vice versa.
\end{theorem}

\begin{example}
Let us look at $ADF_1 = (\{a,b,c\}, \{C_a:\neg a, C_b: a, C_c: \neg b \lor c\}) $, as depicted in Figure \ref{figa}.
$a$ and $b$ cannot form a conflict--free extension to start with, so we are only left with $c$. However, the attack from $b$ on $c$
can be only overpowered by self--support, thus it cannot be part of an aa--admissible extension. Therefore, we obtain only one
aa--preferred extension, namely the empty set.
The single preferred labeling solution would be $v=\{a: \tvu, b:\tvu, c:\tvt\}$ and we can see there is no correspondence between the results.
On the other hand, there is one with the
cc--preferred extension $\{c\}$.

Finally, we have $ADF_2 =(\{a,b,c\}, \{C_a:\neg a \land b, C_b: a,C_c: \neg b\}) $ depicted in Figure \ref{figb}. 
The preferred labeling is $\{a:\tvf, b:\tvf, c:\tvt\}$. The single cc--preferred extension is $\emptyset$ and again, we receive no correspondence.
However, it is compliance with the aa--preferred extension $\{c\}$.

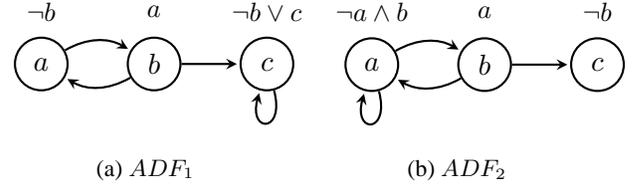
\begin{figure}[t]
\vspace{-1em}
\begin{subfigure}{0.20\textwidth}
  \begin{tikzpicture}
[->,>=stealth,shorten >=1pt,auto,node distance=1.5cm,
  thick,main node/.style={circle,fill=none,draw,minimum size = 0.7cm,font=\large\bfseries},
condition/.style={circle,fill=none,draw=none,minimum size = 0.3cm,font=\normalsize\bfseries}]

\node[main node] (a) {$a$};
\node[main node] (b) [right of=a] {$b$};
\node[main node] (c) [right of=b] {$c$};

\node[condition](ca) [above of= a, yshift=-0.8cm] {$\neg b$};
\node[condition](cb) [above of= b, yshift=-0.8cm] {$a$};
\node[condition](cc) [above of=c, yshift=-0.8cm] {$\neg b \lor c$};
 \path
	(a) edge [bend left] node {} (b)
	(b) edge [bend left] node {} (a)
	(c) edge [loop below] node {} (c)
	(b) edge  node{} (c);
\end{tikzpicture}
\caption{$ADF_1$}
\label{figa}
\end{subfigure}
\quad \,
\begin{subfigure}{0.20\textwidth}
  \begin{tikzpicture}
[->,>=stealth,shorten >=1pt,auto,node distance=1.5cm,
  thick,main node/.style={circle,fill=none,draw,minimum size = 0.7cm,font=\large\bfseries},
condition/.style={circle,fill=none,draw=none,minimum size = 0.3cm,font=\normalsize\bfseries}]

\node[main node] (a) {$a$};
\node[main node] (b) [right of=a] {$b$};
\node[main node] (c) [right of=b] {$c$};

\node[condition](ca) [above of= a, yshift=-0.8cm] {$\neg a \land b$};
\node[condition](cb) [above of= b, yshift=-0.8cm] {$a$};
\node[condition](cc) [above of=c, yshift=-0.8cm] {$\neg b$};
 \path
	(a) edge [loop below] node {} (a)
	(a) edge [bend left] node {} (b)
	(b) edge [bend left] node {} (a)
	(b) edge  node{} (c);
\end{tikzpicture}
\caption{$ADF_2$}
\label{figb}
\end{subfigure}
\caption{Sample ADFs}
\label{counters}\vspace{-1em}
\end{figure}

\end{example}

The labeling--based complete semantics can also be defined in terms of decisiveness:
\begin{theorem}
Let $v$ be a three--valued interpretation and $v'$ its (maximal) two--valued sub--interpretation. $v$ is complete iff
all arguments decisively out w.r.t. $v'$ are mapped to $\tvf$ by $v$ and all arguments decisively in w.r.t. $v'$ are mapped to $\tvt$ by $v$.
\label{thm:complab}
\end{theorem}

Fortunately, just like in the case of admissible semantics, complete extensions and labelings partially correspond:
\begin{theorem}
Let $E$ be a cc--complete and $A$ an aa--complete extension. The $\tvu$--completions of $v_E$ and $v_A^a$ are complete labelings.
\end{theorem}

Please recall that in the Dung setting, extensions and labelings agreed on the sets of accepted arguments. In ADFs, this relation
is often only one way -- like in the case of admissible and complete cc-- and aa-- sub--semantics -- or simply nonexistent, like in
preferred approach. In this context, the labeling--based admissibility (and completeness)
 can be seen as the most general one. This does not mean that specializations, especially handling
cycles, are not needed. Even more so, as to the best of our knowledge no methods for ensuring acyclicity in a three--valued setting are yet available.

Due to the fact that the grounded semantics has a very clear meaning, it is no wonder that both available approaches coincide, as already noted
in \cite{tofix:newadf}. We conclude this section by relating both available notions of stability. The relevant proofs can be found in \cite{report:semantics}.

\begin{theorem}
The two--valued grounded extension and the grounded labeling correspond.
\end{theorem}
\begin{theorem}
A set $M \subseteq S$ of arguments is labeling stable iff it is extension--based stable.
\end{theorem}

\section{Concluding Remarks}

In this paper we have introduced a family of extension--based semantics as well as their classification w.r.t. positive dependency cycles. 
Our results also show that they satisfy ADF versions of Dung's Fundamental Lemma and that appropriate sub--semantics preserve the 
relations between stable, preferred and complete semantics. We have also explained how our formulations relate to the labeling--based approach.
Our results show that the precise correspondence between the extension--based and labeling--based semantics, that holds in the Dung setting,
does not fully carry over. 

It is easy to see that in a certain sense, labelings provide more information than extensions due to distinguishing false and undecided states. Therefore,
one of the aims of our future work is to present the sub--semantics described here also in a labeling form. However, since our focus is primarily on
accepting arguments, 
a comparison w.r.t. information content would not be fully adequate for our purposes and the current characteristic operator could not be fully reused. We hope
that further research will produce satisfactory formulations.

\bibliographystyle{aaai}
\bibliography{references}

\end{document}